\documentclass[11pt]{article}
\usepackage{acl}
\usepackage{amsmath}
\usepackage{amsfonts}
\usepackage{bm}
\usepackage{xspace}             
\usepackage[utf8]{inputenc}     
\usepackage[T1]{fontenc}        
\usepackage{url}                
\usepackage{booktabs}           
\usepackage{amsfonts}           
\usepackage{nicefrac}           
\usepackage{microtype}          
\usepackage{fancyhdr}           
\usepackage{graphicx}           
\usepackage[shortlabels,inline]{enumitem}   
\usepackage{xcolor}             
\usepackage{multirow}           

\usepackage{caption}
\usepackage{subcaption}
\usepackage{wrapfig}
\usepackage{soul}

\usepackage{hyperref}           
\hypersetup{
    pdftitle={PropLogic},
    pdfpagemode=FullScreen,
    }
    
\usepackage{adjustbox}
\usepackage{array}

\usepackage{mdframed}

\usepackage{mathpartir}
\usepackage{proof}

\newcolumntype{R}[2]{%
    >{\adjustbox{angle=#1,lap=\width-(#2)}\bgroup}%
    l%
    <{\egroup}%
}

\newcommand{\smallsection}[1]{\noindent\textbf{#1}.}
\newcommand{\propl}{PropL\@\xspace}
\newcommand{\our}{\textsc{TrialMaster}\@\xspace}

\usepackage{times}
\usepackage{latexsym}
\usepackage[T1]{fontenc}
\usepackage[utf8]{inputenc}
\usepackage{microtype}
\usepackage{inconsolata}
\usepackage{cleveref}
\usepackage{orcidlink}
\usepackage{spverbatim}

\usepackage[utf8]{inputenc}
\usepackage{newunicodechar}
\newunicodechar{∧}{\ensuremath{\land}}
\newunicodechar{∨}{\ensuremath{\lor}}
\newunicodechar{→}{\ensuremath{\to}}

\usepackage{tikz}
\usetikzlibrary{trees, positioning, arrows.meta}

\usepackage{algorithm}
\usepackage{algpseudocode}

\usepackage{listings}
\usepackage{amssymb}
\usepackage{color}
\definecolor{keywordcolor}{rgb}{0.7, 0.1, 0.1}   
\definecolor{tacticcolor}{rgb}{0.0, 0.1, 0.6}    
\definecolor{commentcolor}{rgb}{0.4, 0.4, 0.4}   
\definecolor{symbolcolor}{rgb}{0.0, 0.1, 0.6}    
\definecolor{sortcolor}{rgb}{0.1, 0.5, 0.1}      
\definecolor{attributecolor}{rgb}{0.7, 0.1, 0.1} 

\title{Learn from Failure: Fine-Tuning LLMs with Trial-and-Error Data for Intuitionistic Propositional Logic Proving}

\author{Chenyang An$^1$\thanks{$\ $  The first two authors contributed equally to this work.}, Zhibo Chen$^2$\footnotemark[1]\orcidlink{0000-0003-0045-5024}, Qihao Ye$^1$\orcidlink{0000-0002-7369-757X}, Emily First$^1$, Letian Peng$^1$ \\ \bf Jiayun Zhang$^1$, Zihan Wang$^1$\thanks{$\ $  Corresponding authors.}, Sorin Lerner$^1$\footnotemark[2], Jingbo Shang$^1$\footnotemark[2] \\
University of California, San Diego$^1$ \quad Carnegie Mellon University$^2$ \\
  \texttt{\{c5an, q8ye, emfirst, lepeng, jiz069, ziw224, lerner, jshang\}@ucsd.edu}\\
  \texttt{zhiboc@andrew.cmu.edu} 
  }

\begin{document}
\maketitle
\begin{abstract}
Recent advances in Automated Theorem Proving have shown the effectiveness of leveraging a (large) language model that generates tactics (i.e. proof steps) to search through proof states.
The current model, while trained solely on successful proof paths, faces a discrepancy at the inference stage, as it must sample and try various tactics at each proof state until finding success, unlike its training which does not incorporate learning from failed attempts.
Intuitively, a tactic that leads to a failed search path would indicate that similar tactics should receive less attention during the following trials. 
In this paper, we demonstrate the benefit of training models that additionally learn from failed search paths.
Facing the lack of such trial-and-error data in existing open-source theorem-proving datasets, we curate a dataset on intuitionistic propositional logic theorems and formalize it in Lean, such that we can reliably check the correctness of proofs. 
We compare our model trained on relatively short trial-and-error information (\our) with models trained only on the correct paths and discover that the former solves more unseen theorems with lower trial searches.
\end{abstract}

\section{Introduction}

Automated Theorem Proving is a challenging task that has recently gained popularity in the machine-learning community. Researchers build \emph{neural theorem provers} to synthesize formal proofs of mathematical theorems~\cite{yang2023leandojoa,welleck2021naturalproofs,lample2022hypertree,mikula2023magnushammer,wang2023dt,bansal2019learning,davies2021advancing,wu2021tacticzero,rabe2020mathematical,kusumoto2018automated,bansal2019learning,irving2016deepmath}.
Typically, a neural theorem prover, given a partial proof and the current \emph{proof state}, uses a neural model to predict the next likely \emph{proof step}, or \emph{tactics}. The neural models utilize different architectures like LSTMs~\cite{sekiyama2017towards}, CNNs~\cite{irving2016deepmath}, DNNs~\cite{sekiyama2018automated}, GNNs~\cite{bansal2019learning,wang2017premise} and RNNs~\cite{wang2020learning}, though most recent work has begun to explore the use of transformer-based large language models (LLMs) due to their emerging reasoning abilities. 

\begin{figure}[t]
    \centering
    \includegraphics[width=\linewidth]{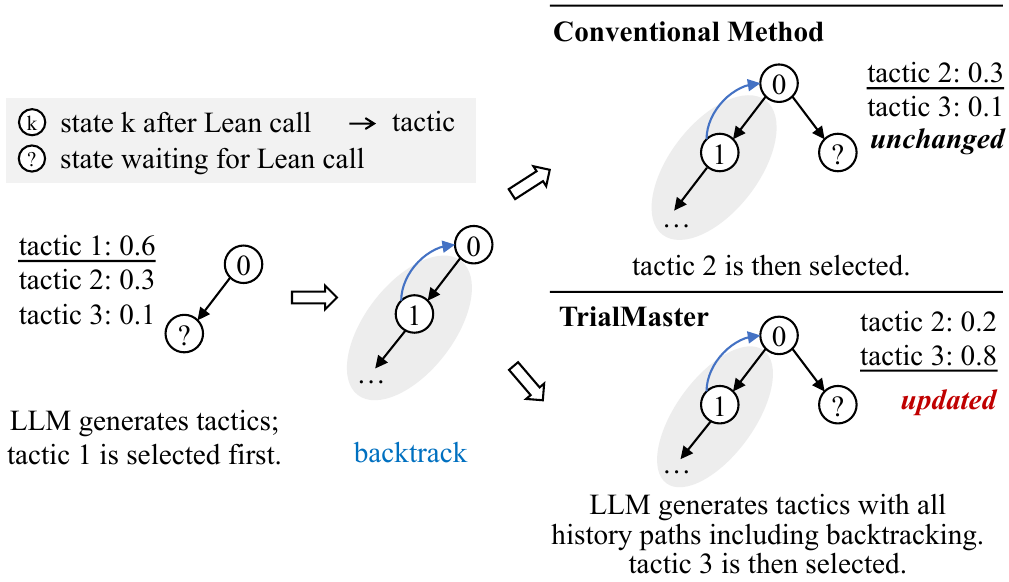}
    \caption{A simple example for how learning trial-and-error data impacts inference distribution. See Figure ~\ref{fig:intro-propl-example} for a concrete case.}
    \label{fig:trial-and-error-inference}
    \vspace{-2mm}
\end{figure}

\begin{figure*}[t]
    \centering
    \begin{subfigure}{0.48\textwidth}
        \centering
        \includegraphics[width=\linewidth]{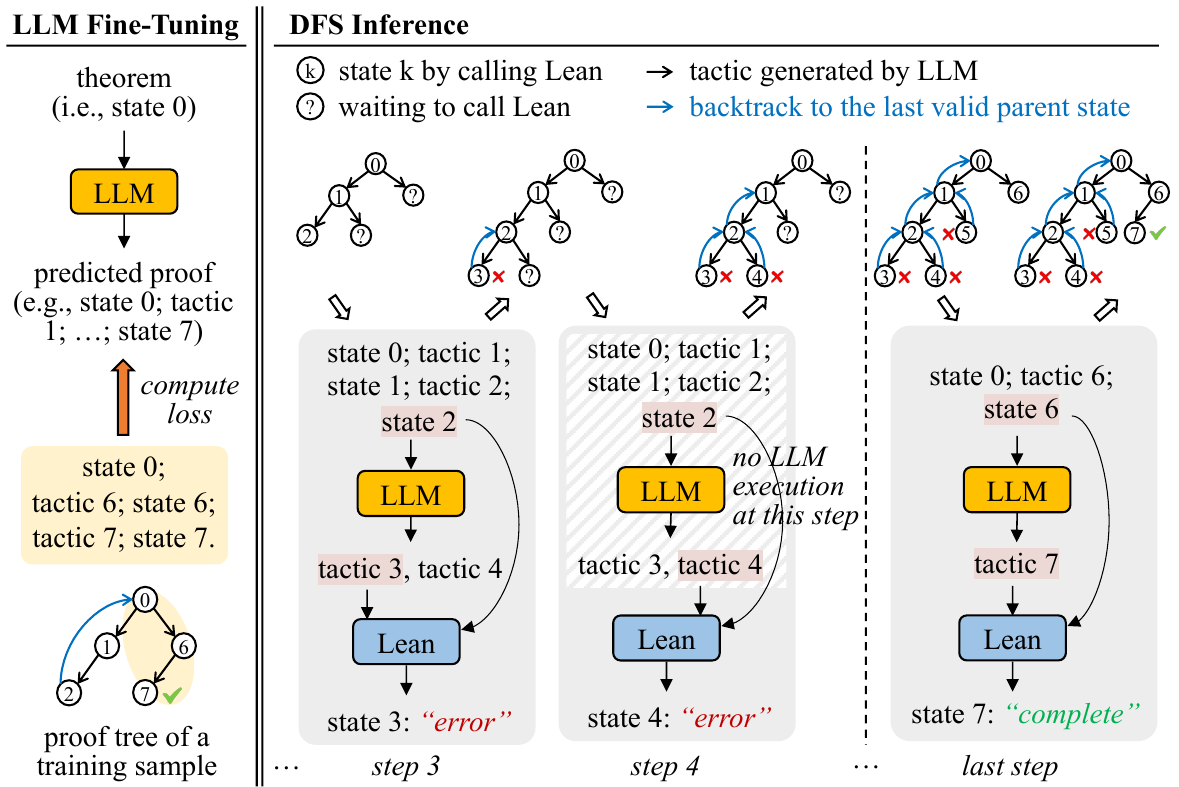}
        \caption{Conventional system with depth-first search}
        \label{fig:method-dfs}
    \end{subfigure}
    \hfill
    \begin{subfigure}{0.48\textwidth}
        \centering
        \includegraphics[width=\linewidth]{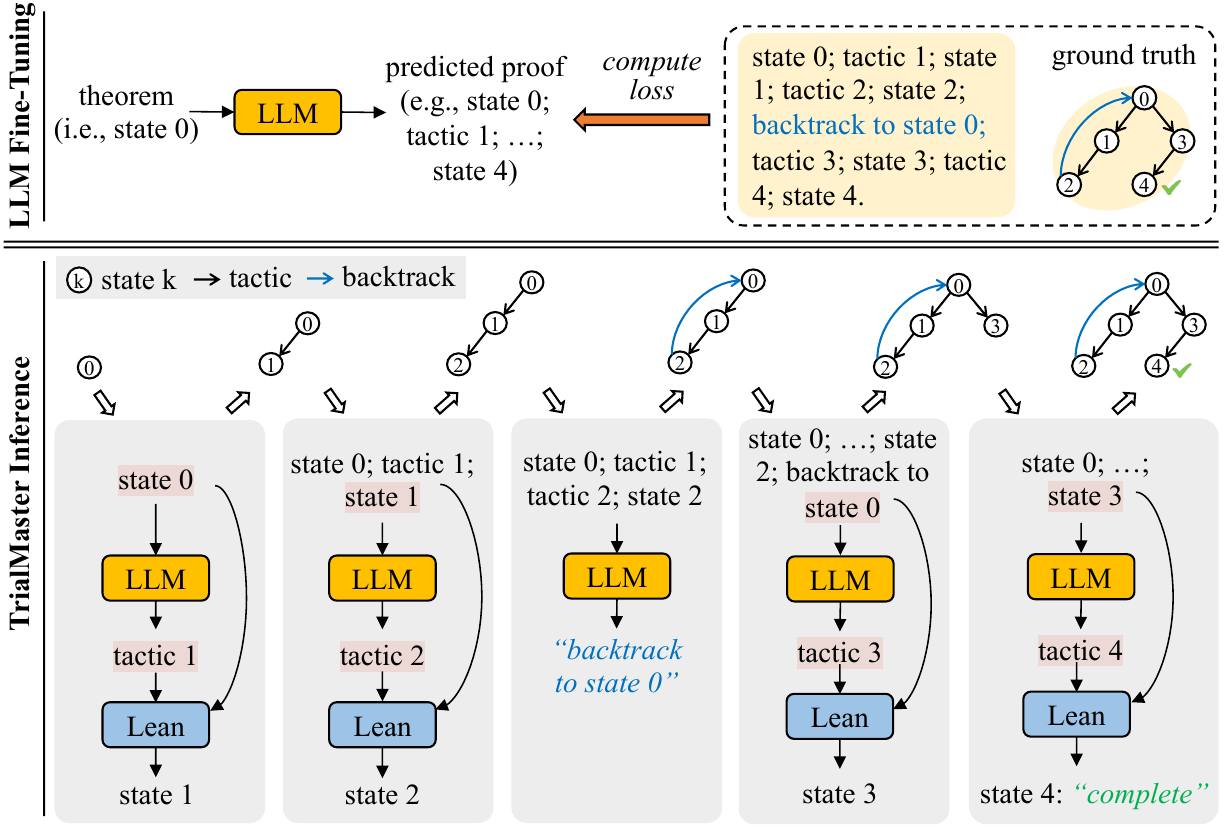}
        \caption{Our training and inference methodology}
        \label{fig:method-ours}
    \end{subfigure}
    \caption{Method comparison. (a) A conventional system: The tactic generator (i.e., LLM) is fine-tuned on correct proof paths only. During inference, the trained tactic generator produces $N_{\text{sampled}}$ (e.g., 2 in the example) tactics at a time. 
    If Lean decides that the current tactic is wrong, the system backtracks to the last valid state and tries other candidate tactics. (b) Our methodology: The tactic generator is fine-tuned on proofs with trial-and-error. During inference, we take the first tactic it generates and feed that into Lean for state checking at each step.
    }
\end{figure*}

An interactive \emph{proof assistant}, such as Lean~\cite{de2015lean}, Coq~\cite{barras1997coq} or Isabelle~\cite{nipkow2002isabelle}, evaluates the model's predicted candidate proof steps, returning either new proof states or errors.
Neural theorem provers iterate on this procedure, performing \emph{proof search}, e.g., a depth-first search (DFS), to traverse the space of possible proofs. An example of a DFS proof search is illustrated in Figure~\ref{fig:method-dfs}, where the prover progressively generates new tactics if the attempted tactics result in incorrect proofs.

Such provers are usually trained on a dataset containing only the correct proof paths.
This, however, presents a limitation: during inference, the prover does not have the ability to leverage the already failed paths it explored. 
Such failure information, intuitively, is beneficial, as it could suggest the model to generate tactics similar to the failed ones sparingly. 
At the very least, the failure information should help the model easily avoid generating already failed tactics. See Figure~\ref{fig:trial-and-error-inference}.

In this paper, we wish to empirically verify this intuition.
To conduct the experiment, we would compare the conventional model trained on correct proof paths, and \our, the model trained on the whole proof tree, containing both correct paths and incorrect paths. See Figure ~\ref{fig:method-ours}.
As such, \our can make predictions based on the failure information during inference time.

Since current open-source Automated Theorem Proving datasets do not contain complete proof trees, we create such a dataset, \propl, written in Lean. 
We focus on theorems of intuitionistic propositional logic. A propositional logic formula in intuitionistic propositional logic $A$ is true if and only if $\vdash A$ has a derivation according to the rules of the intuitionistic propositional logic. These rules are listed in Figure~\ref{fig:ipl-rules} in \Cref{sec:IRIPL} with explanations. This is to be distinguished from the classical logic, where a theorem of classical propositional logic admits a proof if and only if under all truth value assignments to all propositions that appear in the theorem, the theorem statement is evaluated to be true.

We give two elementary examples of theorems of intuitionistic propositional logic with proofs. We then give an example of a theorem whose proof contains a backtracking step. The first theorem admits a proof, where as the second doesn't.
\begin{lstlisting}
     theorem_1 thm : p1 &$\rightarrow$& p1 $\vee$ p2 := by
         intro h1
         apply or.inl
         exact h1
\end{lstlisting}

The first theorem states that $\texttt{p1}$ implies $\texttt{p1} \lor \texttt{p2}$, where $\lor$ stands for "or". To show this, we assume $\texttt{p1}$ and prove either $\texttt{p1}$ or $\texttt{p2}$. We prove $\texttt{p1}$ in this case. The last three lines are tactics (proof steps) in Lean used to prove this fact.

Our second example looks like the following.
\begin{lstlisting}
     theorem_2 thm : p1 &$\rightarrow$& p1 $\land$ p2 := by
\end{lstlisting}

The second theorem states that $\texttt{p1}$ implies $\texttt{p1}$ and $\texttt{p2}$. There is no proof to this theorem. When we assume $\texttt{p1}$, we cannot show both $\texttt{p1}$ and $\texttt{p2}$.

Proofs might include backtracking instructions. Here is another possible proof for theorem 1. After step 2, there is no possible proof step that can lead to final solution. Therefore we backtrack to step 1 and try a different proof step.

{
\lstset{
    commentstyle=\color{gray}, 
    morecomment=[l]{\#}, 
}

\begin{lstlisting}
     theorem_1 thm : p1 &$\rightarrow$& p1 $\vee$ p2 := by
         intro h1 #this is step 1
         apply or.inr #this is step 2
         no solution, backtrack to step 1
         apply or.inl
         exact h1
\end{lstlisting}
}

Note that "no solution, backtrack to step 1" is not a tactic in Lean. It simply tells the system to backtrack to step 2 and start reasoning from there.

Specifically, our \propl dataset is created through a two-stage process that first involves generating a comprehensive set of propositional logic theorems by uniformly sampling from all possible theorems, utilizing a bijection between natural numbers and propositions to ensure representativeness. Following theorem generation, proofs are constructed using a focusing method with polarization, incorporating a detailed trial-and-error search process that includes both successful and backtracked steps, thereby capturing the complexity and nuances of theorem proving in intuitionistic propositional logic. Thus, our dataset is complete, scalable, and representative. The proofs in our dataset are combined with trial-and-error information, which is generated by the Focused Proof Search (FPS) algorithm~\cite{mclaughlin09efficient, liang09focusing, pfenning17lnfocusing}. 

We verify the effectiveness of incorporating the failed trials during training and inference by experiments on \propl, observing that \our achieves a higher proof search success rate and lower search cost over conventional model trained on correct proof paths. Our experiments further indicate that our model can perform backtracking without help from an external system.

Our main contributions are as follows:
\begin{itemize}[nosep,leftmargin=*]
    \item We establish, \propl, a complete, scalable, and representative benchmark for intuitionistic propositional logic theorems formalized in Lean. \propl includes proofs with trial-and-error information, generated by the FPS algorithm. 
    
    \item We demonstrate that for intuitionistic propositional logic theorem proving, incorporating trial-and-error information into training and proving outperforms a conventional model that is trained on correct proofs only.
\end{itemize}

\section{Related Work}
\smallsection{Automated Theorem Proving} 
Automated Theorem Proving has evolved significantly since its inception, focusing mainly on developing computer programs that can autonomously prove mathematical theorems.
Early ATP systems (mechanical theorem proving) were based on first-order logic~\cite{clocksin2003programming,chang2014symbolic}, where the resolution method~\cite{robinson1965machine} played a crucial role. 
Recent progress in Automated Theorem Proving has been marked by the integration of machine learning~\cite{bansal2019learning,davies2021advancing,wagner2021constructions}, especially LLMs~\cite{yang2023leandojoa,polu2020generative,han2021proof,welleck2021naturalproofs,jiang2022draft}, and heuristic methods~\cite{holden2021heterogeneous}, aimed at amplifying the efficiency and capacity of Automated Theorem Proving systems.
Within the domain of LLMs, formal mathematical languages like Metamath~\cite{megill2019metamath}, Lean~\cite{de2015lean}, Isabelle~\cite{nipkow2002isabelle}, and Coq~\cite{barras1997coq}, serve as a bridge, enabling the precise expression and verification of mathematical theorems and concepts through a computer-verifiable format, thereby mitigating hallucination risks~\cite{nawaz2019survey}. COPRA incorporates backtracking information into a prompt and sends the prompt to GPT-4 without fine-tuning it to perform proof search~\cite{thakur2023language}. Baldur fine-tunes LLMs with proofs and error information given by the proof assistant~\cite{first2023baldur}. In contrast, our work focuses on fine-tuning LLMs with the complete past proof history without using the error message from the proof assistant.

\smallsection{Propositional Logic Problem}
Early implementations of ATP systems demonstrated the potential for computers to automate logical deductions, with notable examples including the Logic Theorist \cite{crevier1993ai,McCorduck2004-MCCMWT,russell2010artificial} and Gilmore's program \cite{davis2001early,5392528}.
These systems laid the groundwork for the resolution of propositional logic problems, showcasing the ability of automated systems to handle logical reasoning tasks.
Recent advancements in Automated Theorem Proving have revisited propositional logic problems, integrating modern computational techniques. 
Sekiyama et al.~\cite{sekiyama2018automated} have employed Deep Neural Networks (DNNs) as a statistical approach to generate proofs for these theorems, while Kusumoto et al.~\cite{kusumoto2018automated} have explored graph representations coupled with reinforcement learning to find proofs.
Furthermore, sequence-to-sequence neural networks have been applied for deriving proof terms in intuitionistic propositional logic~\cite{sekiyama2017towards}.
This area of research is particularly intriguing due to the simplicity and importance of propositional logic in mathematics, and there is a growing interest in evaluating the capability of LLMs in tackling this specific mathematical domain.

\smallsection{Trial-and-Error}
The Chain-of-Thought (CoT)~\cite{wei2022chain,wang2022self,zhou2022least,fu2022complexity,chu2023survey,yu2023nature} approach demonstrates that LLMs can be guided to perform step-by-step reasoning by incorporating intermediate reasoning steps in their prompts.
This concept is expanded in later research, such as the Tree of Thoughts (ToT)~\cite{yao2023tree}, which organizes reasoning into a tree structure, and the Graph of Thoughts (GoT)~\cite{besta2023graph}, which adopts a graph format for thought structuring.
Trial-and-error complements structured reasoning by allowing the model to empirically test hypotheses generated, thereby refining its reasoning process based on feedback from interactions or emulations.
The Boosting of Thoughts (BoT)~\cite{chen2024boosting} prompting framework iteratively explores and evaluates multiple trees of thoughts to gain trial-and-error reasoning experiences, using error analysis from the LLMs to revise the prompt.
Lyra introduces correction mechanisms to improve the performance of the tactic generator~\cite{zheng2023lyra}.

\section{\propl: A New Dataset for Intuitionistic Propositional Logic Theorems in Lean} 
\label{sec:data}

Our aim is to experimentally validate that trial-and-error information can enhance models' ability to do backtracking and tactic generation for theorem-proving tasks. Given that existing open-source theorem proving datasets lack information on trial-and-error processes, we have developed \propl, which is based on theorems of intuitionistic propositional logic. This dataset uniquely includes proofs that encapsulate the complete search process, incorporating the trial-and-error data generated by the FPS algorithm. Our dataset has two other benefits. It is formalized in Lean, so that the validity of the theorems and proofs are guaranteed. The tactics generated by the model trained on \propl can also be directly sent to Lean to be checked. \propl is also representative of all the intuitionistic propositional logic theorems, since by uniformly sampling integers, we can use a bijection between natural numbers and propositions to uniformly sample propositions. This bijection is explained in the Theorem Generation section.
\subsection{Data Generation of \propl}
\propl comprises theorems uniformly sampled from the entire set of propositional logic theorems. It includes various proof types for each theorem. We only report proof types that are used in this paper. For additional information about the dataset, please refer to the GitHub repository and Huggingface.

The construction of \propl involves two primary stages: the generation of propositional logic theorems and the generation of proofs for these theorems from an existing algorithm.

\begin{figure}[t]
    \centering
    \includegraphics[width=0.75\linewidth]{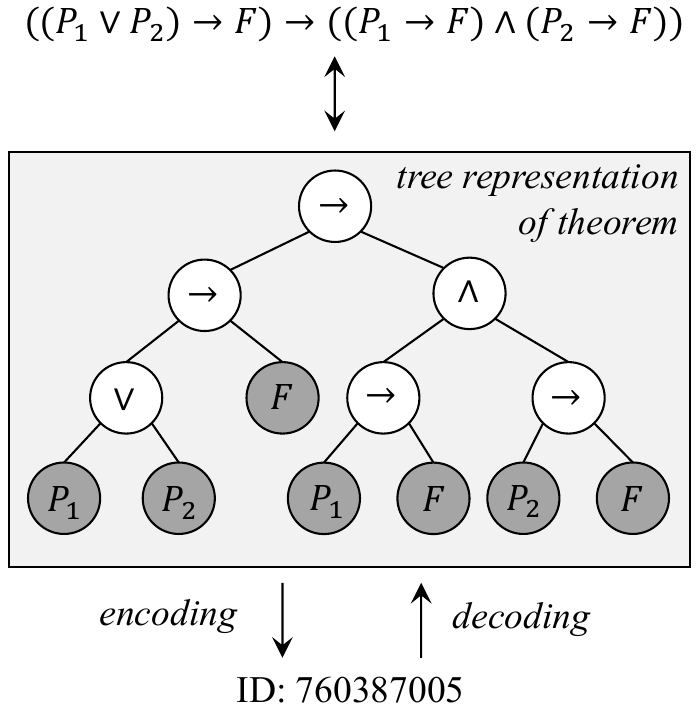}
    \caption{An illustration of the bijection between a proposition and a natural number, where gray nodes are leaf nodes. ID is computed using \Cref{alg:encoding} with $n = 6$ and $p = 2$ in this case.}
    \label{fig:proposition_number_mapping}
\end{figure}

\smallsection{Theorem Generation}
Consider the set of propositions $A$ with at most $p$ atomic propositions. $A$ can be inductively generated by the following grammar:
\begin{equation*}
    A, B
    ::= P_{i} \mid T \mid F \mid A \land B \mid A \lor B \mid A \to B,
\end{equation*}
where $P_i$ is the $i$-th atomic proposition with $1 \le i \le p$, $T$ stands for True, and $F$ stands for False. We note that an atomic proposition is just a statement like "Today is sunny". The concrete meaning of the statement is irrelevant with respect to the theorem proving task in this paper.
A connective $\land$, $\lor$, $\to$ is called an internal node of the proposition.
We assign the following lexicographic order to propositions:
\begin{enumerate}[nosep]
    \item Number of internal nodes (increasing order)
    \item Number of internal nodes of the left child (decreasing order)
    \item Top level connective, $T < F < P_{1} < \cdots < P_{p} < \land < \lor < \to$
    \item The (recursive order (2 - 5)) of the left child
    \item The (recursive order (2 - 5)) of the right child
\end{enumerate}
For a fixed upper bound $p$ for the number of atomic propositions, we establish a bijection between the natural numbers and the set of propositional logic formulas. The counting can be made efficient using the Catalan Numbers~\cite{atkinson1992generating}. 
\Cref{fig:proposition_number_mapping} gives an example of mapping between propositions and natural numbers. Details about the encoding and decoding algorithms are provided in \Cref{sec:UDDE}.

\begin{figure}
    \centering
    \includegraphics[width=\linewidth]{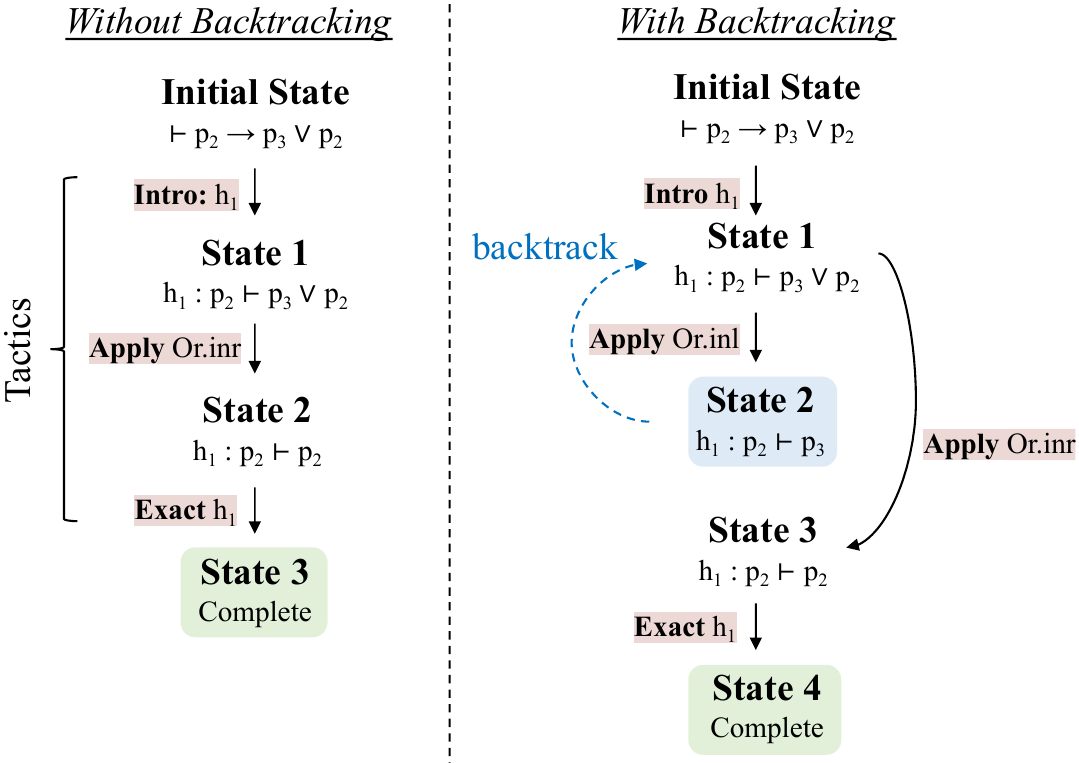}
    \caption{Two proofs for one propositional logic theorem with tactics and states in Lean.
    }
    \label{fig:intro-propl-example}
\end{figure}

\smallsection{Proof Generation}
Given a randomly sampled theorem, the proof of
the theorem is constructed using the focusing method with polarization \cite{mclaughlin09efficient, liang09focusing, pfenning17lnfocusing}. Proof search is divided into two stages: inversion and chaining. The inversion phase mechanically breaks down negative connectives (e.g. implications) in the goal and positive connectives (e.g. disjunctions) in the premises. After inversion, chaining will pick an implication in the premise or show one of the disjuncts in the conclusion, with backtracking. The proof search procedure terminates when the same atomic proposition appears in both the premise and the conclusion. An example of proofs with trial-and-error information (backtracking) and with trial-and-error information removed are shown in Figure~\ref{fig:intro-propl-example}.

Once proofs are generated, we use them to fine-tune models and start the proof search on the test set. 

The polarization of the connectives affects the behavior of the inversion and the search procedure. We choose to uniformly polarize conjunctions that occur negatively (e.g. on the right-hand side of an arrow)  as negative and polarize conjunctions that occur positively (e.g. on the left-hand side of an arrow) as positive. Atomic propositions are assigned polarities based on the connective that they first appear under.

To improve the runtime of the search procedure, we make an additional assumption that once an implication is picked, the implication cannot be used to show its premise. In theory, this introduces incompleteness into the search procedure, but it only affects 1 theorem out of around 1000 provable theorems randomly sampled.

\subsection{Construction of Training and Test Sets}

In this section, we explain how we construct the datasets for training and evaluation. We want to avoid training and testing on similar data. In order to test the model performance on harder out-of-distribution (OOD) tasks, we need to ensure that the lengths of the proofs in the training data are shorter than the lengths of the proofs in the test data.

Given \propl, we fix the number of internal nodes in the theorem statement to be 16 (explained in the dataset generation section). We then uniformly randomly sample 200,000 theorems from \propl, which can be achieved by using the integer-theorem bijection as explained before. Our method ensures that the theorems we study in this paper are representative of the propositional logic theorems in general.

We first apply our deterministic algorithms to generate the proofs of the 200,000 theorems, and then remove the trial-and-error information of those proofs. We get a word-length distribution of those proofs without trial-and-error information.
Next, to ensure the diversity of the trial-and-error information, we randomly select propositions to focus on during the chaining phase of the proof search, and then generate 10 different proofs with backtracking. By using the average length of the 10 different proofs, we have another word length distribution of proofs with trial-and-error information.

We then split the 200,000 theorems into training and test sets based on both of the word length distributions mentioned above. The word lengths of the proofs of the training data theorems fall within the lower 0.66 quantile of the two distributions of the word length of all the proofs of the 200,000 theorems (109,887 in total). The word lengths of the in-distribution test set also fall in that category (1000 in total). The word lengths of the proofs among the out-of-distribution test theorems are above 0.8 quantile (1000 total) of the two distributions of the word lengths of all the proofs of the 200,000 theorems. We note that the baseline model and \our are trained on proofs of the same set of theorems.

\section{Methodology}

\subsection{LLM Fine-Tuning}
We utilize the training set formed in \propl for training both \our and the tactic generator in the DFS system. The numbers of theorems used in the training and test datasets are presented in Table~\ref{tab:train-test-split}. 

\smallsection{LLM fine-tuning with trial-and-error}
In our approach, we randomly select two out of the shortest five among the ten proofs with trial-and-error information for each theorem in the training set and utilize them to train \our. Refer to Figure~\ref{fig:method-ours} for a description of this training process.

\smallsection{LLM fine-tuning in a DFS system}
For the tactic generator of the DFS system, we employ the deterministic FPS algorithm to generate the proofs of the theorems in the training set. The trial-and-error information is removed from the proofs. The LLM is then fine-tuned on the proofs without trial-and-error information as the conventional methods do. Figure~\ref{fig:method-dfs} illustrates the training process of the DFS system.

\begin{table}[t]
\centering
\caption{The scale of training and test split in our \propl dataset.}
\label{tab:train-test-split}
\begin{tabular}{lcc}
\toprule
Subset       &  Number of theorems   \\ 
\midrule
Training set     &  109,887   \\
In-dist. test set    & 1,000     \\
Out-of-dist. test set   & 1,000    \\
\bottomrule
\end{tabular}
\end{table}

\subsection{Inference}
\smallsection{Inference method of model trained with trial-and-error}
\our conducts inference on itself without any help from a backtracking system like DFS or BFS. It outputs two kinds of tactics: tactics in Lean and backtrack instructions. An example of a backtrack instruction would be like \textit{``no solution, return to state 2 [that leads to state 4]''}, where state 4 is the current state. When \our is doing a proof search on the test set, it is prompted with all history paths, including previous tactics, states, the backtracking it made before, and the failed search path. It then outputs the entire proof path after. Nonetheless, we only utilize the first tactic in the output and employ Lean as a calculator to determine the next state, thereby ensuring the correctness of the state following the tactic. If the tactic outputted by \our is a backtrack instruction, it is then prompted with all the proof search history including the backtrack instruction and the state that the backtrack instruction says to return to. If that tactic is not a backtrack instruction, the tactic and the current state will be fed into Lean for producing the state after. \our is then prompted with the entire proof tree including the state that Lean calculated, and it should output a tactic again. This process is repeated until Lean identifies that the proof is complete or any Lean error occurs. We also note that \our only outputs one tactic at each state using greedy search.

\smallsection{Inference method of the DFS system}
There are two hyperparameters in the DFS system: temperature $t$ and the number of sampled tactics $N_{\text{sampled}}$. The temperature $t$ decides the diversity of the model outputs. As $t$ increases, the outputs of the model become more varied. The second parameter determines how many tactics the tactic generator of the DFS system produces for a new proof state.

During inference, the LLM in the DFS system produces $N_{\text{sampled}}$ of candidate tactics at each new state. For each proof state, the DFS only makes one inference. If any two of the generated tactics for the same state are the same, we remove one of them to ensure efficiency. We also remove the tactic suggestions that fail to follow the grammar of Lean. The system follows the depth-first order to keep trying untried tactics. If the system exhausts all the tactics for a given state but has not found a valid one, the system returns to the parent state and then keeps trying untried tactics for the parent state. 
The overview is presented in the \Cref{fig:method-dfs}. 

To fully exploit the ability of the DFS system, we varied the parameters of it, such as temperature and the number of sampled tactics.  We count how many times Lean has been called to check tactics for both the DFS system and \our during the inference stage.

\paragraph{Why do we choose DFS over BFS?} While the breadth-first-search (BFS) system is also popular for building neural provers in Automated Theorem Proving, we have opted for DFS as our baseline over BFS in the context of propositional logic theorem proving. This is due to the finite number (around 20) of tactics available at any step for the search process of intuitionistic propositional logic theorems, making DFS more efficient than BFS without compromising the success rate.

\section{Evaluation}

\subsection{Experimental Setup}
\smallsection{Base LLM} We used \texttt{Llama-2-7b-hf} \cite{touvron2023llama} as the backbone LLM for tactic generation. Models are trained on two A100 GPUs for a single epoch with batch size set to $4$. 
Huggingface \cite{wolf2019huggingface} is used for fine-tuning the models, and VLLM~\cite{kwon2023efficient} is used for inference of the models for efficiency. The learning rate of all training processes is set to be $2\times 10^{-5}$. 

\smallsection{Hyperparameters}\label{sec:experiment-setup}
In the experiment, we vary temperature $t$ from $\{0.3, 0.7, 1.2, 1.5, 2.0\}$ and number of sampled tactics $N_{\text{sampled}}$ from $\{2, 5, 10, 15, 20\}$. We notice that in all experiments for temperature $t$ less than 1.5, there were only very few (less than 15) theorems that were still under search when the total steps for that search attempt reached 65. For $t$ higher than 1.5, $N_{\text{Lean}}$ starts to increase dramatically. At temperature 2 with $N_{\text{sampled}}$=20, $N_{\text{Lean}}$ climbs up to 32,171 when the number of total steps reaches 65. Therefore, in our experiment, we set 65 as the search step limit to control time complexity.

\begin{figure*}[t]
    \centering
    \begin{subfigure}{0.31\linewidth}
        \centering
        \includegraphics[width=\linewidth]{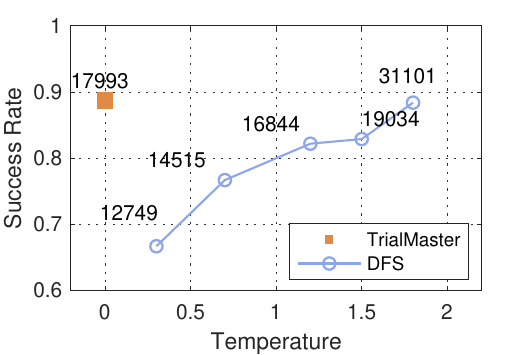}
        \caption{Temperature $t$}
        \label{fig:eval-temp}
    \end{subfigure}
    \hfill
    \begin{subfigure}{0.31\linewidth}
        \centering
        \includegraphics[width=\linewidth]{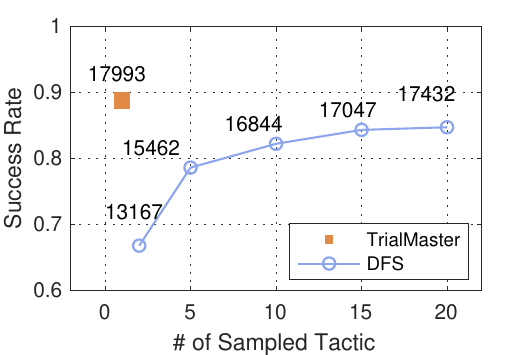}
        \caption{Sampled Tactics $N_{\text{sampled}}$}
        \label{fig:eval-NST}
    \end{subfigure}
    \hfill
    \begin{subfigure}{0.31\textwidth}
        \centering
        \includegraphics[width=\linewidth]{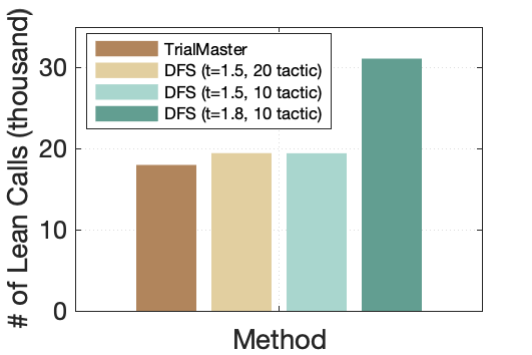}
        \caption{Search Cost}
        \label{fig:eval-lean}
    \end{subfigure}
    \caption{Experiment results on \textbf{OOD} task. (a) We fix $N_{\text{sampled}} = 10$ to see the impact of temperature on the DFS system. (b) We fix $t=1.2$ to see the impact of the number of sampled tactics. The number of Lean calls is noted beside the marker. (c) Comparison of $N_{\text{Lean}}$ among our method and top 3 DFS systems with the highest success rate. In summary, training with trial-and-error achieves a higher success rate with a relatively lower search cost compared to the DFS systems.}
    \label{fig:eval-NST-temp}
\end{figure*}

\subsection{Evaluation Metrics}
\smallsection{Proof search success rate} We use proof search success rate as our primary metric, which represents the ratio of successful searches by the model. 
A search attempt for a theorem is marked as successful if Lean outputs state \textit{``no goal, the proof is complete''}, implying the theorem is effectively proved after applying a tactic produced by the model.
For \our, a search attempt ends and fails immediately after one of the following conditions: 1) the word length of the proof tree with trial-and-error exceeds $1500$ (for the sake of context length of the model), or 2) the tactic produced by the model at any step induces a Lean error.
For the conventional DFS system, a search attempt fails when one of the following conditions happens: 1) the word length of the proof tree without trial-and-error exceeds 1500, or 2) all tactics generated for the initial states have been explored and failed, or 3) the total search steps exceed 65 (see Section~\ref{sec:experiment-setup} for the choice of this value).
We note that the 1500-word limit is stricter for our method since it produces entire proof paths including trial-and-error and thus would be easier to hit the limit.

\smallsection{Search cost}  We define a metric to assess the search cost---the total number of Lean calls for tactic checking during proof search for the entire test set, denoted as $N_{\text{Lean}}$. Given the same proof search success rate, a lower $N_{\text{Lean}}$ indicates a more efficient system for proof search. 
Note that backtracking instructions from our method do not require applying Lean to check the state and, consequently do not add search cost. 

\begin{table}[t]
  \centering
  \small
  \caption{Performance on \textbf{in-distribution} task. Both methods perform well for propositional logic.}
  \label{model-in-distribution}
    \begin{tabular}{cccc}
    \toprule
    \multicolumn{3}{c}{Model} & \multicolumn{1}{l}{Success Rate} \\
    \midrule
    \multicolumn{3}{c}{\our} & \textbf{100\%} \\
    \midrule
    \multirow{6}[3]{*}{DFS} & \multirow{3}[2]{*}{$t=1.2$} & $N_\text{sampled}=2$   & 99.5\% \\
          &       & $N_\text{sampled}=5$   & 99.9\% \\
          &       & $N_\text{sampled}=10$  & 99.6\% \\
\cmidrule{2-4}          & \multirow{3}[2]{*}{$t=2.0$} & $N_\text{sampled}=2$   & 75.9\% \\
          &       & $N_\text{sampled}=5$   & 97.3\% \\
          &       & $N_\text{sampled}=10$  & 99.0\% \\
    \bottomrule
    \end{tabular}%
\end{table}%

\subsection{Results and Analysis} 

\smallsection{\our outperforms conventional DFS system}
We begin by evaluating the methods of the in-distribution test set. Table~\ref{model-in-distribution} illustrates that both our method and the DFS system perform exceptionally well, achieving a success rate of nearly 100\% in most configurations. This suggests that Llama-7b effectively masters in-distribution intuitionistic propositional logic theorems. Then, we compare the performance of the methods on the out-of-distribution task.
The results are presented in \Cref{fig:eval-NST-temp}. 
Our method with trial-and-error significantly outperforms the DFS system across various hyperparameter configurations.  
Additionally, we observe that feeding more proofs without trial-and-error for LLM fine-tuning does not further improve the performance.

\smallsection{Impact of hyperparameters in the DFS system}
As shown in Figure~\ref{fig:eval-NST-temp}, on the OOD task, although the success rate of the DFS system gets higher when we increase the temperature $t$ or the number of sampled tactics $N_{\text{sampled}}$, the search cost (reflected by $N_\text{Lean}$) also goes up. 
Specifically, when we increase $N_{\text{sampled}}$, the DFS system explores a larger pool of candidate tactics during the search, leading to a higher number of Lean calls.
In contrast, our method does a greedy search to generate only one tactic for each new state.
Likewise, as $t$ increases, the tactic generator of the DFS system tends to produce more diverse tactics at each proof state, improving the system's performance but leading to higher search costs.

\begin{table}[t]
\centering
\caption{Ablation study: Comparison of \our and model trained without trial-and-error information on OOD task}
\label{two-model-comparison-wo}
\begin{tabular}{lcc}
\toprule
Model       &  Success rate   \\ 
\midrule
\our     &  \textbf{88.7\%}   \\
Model - proof w/o t.a.e.  & 59.3 \%    \\
\bottomrule
\end{tabular}
\end{table}

\smallsection{\our achieves high success rates at lower search cost}
For a direct comparison of search costs, we plot the $N_\text{Lean}$ values of our method alongside those of the top three DFS systems with the highest success rates among all the hyperparameters we experimented with, i.e., $t$=1.5, $N_\text{sampled}$=20 (87.2\%), $t$=1.5, $N_\text{sampled}$ = 10 (86.1\%), and $t$=1.8, $N_\text{sampled}$=10 (88.4\%). This comparison is illustrated in Figure~\ref{fig:eval-lean}. Notably, we observe that the DFS systems that closely approach our model's performance exhibit significantly higher search costs. With $t$=1.8, $N_\text{sampled}$=10, $N_{\text{Lean}}$ of the DFS system, which has 0.3\% lower success rate than our method, has reached 31,101, which is 72\% higher than that of our method with trial-and-error. 
The high search cost makes the DFS system with high temperatures unfavorable.
These results demonstrate that training with trial-and-error produces higher-quality tactics, achieving a higher success rate with relatively lower search cost. 

\smallsection{Model learns backtracking capability from trial-and-error data}
In the experiments, we find out that our \our successfully acquires the backtracking capability from proofs with trial-and-error information. This is evidenced by the fact that during \our's proof search for theorems in the test set, all backtracking instructions produced by the LLM adhere to the correct format and point to existing state numbers.

\smallsection{Including failed search paths helps \our to learn}
The following experiment shows that adding failed search paths to the training data for \our results in an overall gain. In this experiment, the model is only trained to learn the correct search paths and the backtracking instructions. The model is not trained to learn the failed search paths (we don't compute the loss for the failed search paths during the training stage in this case). The proof success rate in this case is 75.6\%, which is lower than \our's proof success rate of 88.7\%. The $N_{\text{Lean}}$ for the model is 13,600, which is lower than that of \our. This is expected since the model does not learn to predict failed search paths. Our explanation for why \our has a higher proof search success rate than the model trained in the previously mentioned experiment is that the failed search paths also contribute to improving the proof search success rate of the model. \our strategically tried some potentially failed search paths to gain a more comprehensive view of the problem, which then led to the correct search paths.

\subsection{Ablation Study}

To evaluate the effectiveness of training with trial-and-error, we craft an ablated version of our method where the LLM is fined-tuned with data of the correct path only and do inference in the same way as our method (i.e., producing one tactic at a time and applying Lean for state checking). We denote the ablated version as \textit{Model - proof w/o t.a.e.}. For both methods, we mark the search attempt as failed if the tactic induces a Lean error, or the search exceeds the 1500-word limit.
The result is shown in the Table~\ref{two-model-comparison-wo}.
The difference between the success rates of the two models is $29.4\%$, which is significant. This clearly shows that failed search states and trial-and-error information tremendously enhance the model's capability to solve theorem-proving tasks.

\begin{table}[t]
\centering
\caption{Comparison of models trained on different lengths of proofs with trial-and-error on OOD task.}
\label{three-model-comparison}
\begin{tabular}{lcc}
\toprule
Model       &  Success rate   \\ 
\midrule
Model - short proof w/ t.a.e.      &  \textbf{88.7\%}   \\
Model - long proof w/ t.a.e.    & 72.4 \%    \\
\bottomrule
\end{tabular}
\end{table}

\subsection{Exploratory Study: Effect of Training Proof Length on Model Performance}
Since the FPS algorithm of \propl dataset can generate multiple proofs with variable length, we conduct an exploratory study to assess the impact of proof length on model performance.
We fine-tune two models using proofs with different lengths of trial-and-error information. 
For the first model, which is our \our, the training data is derived by randomly selecting two out of the shortest four proofs from the ten available proofs for each theorem in \propl. We denote it as \textit{Model - short proof w/ t.a.e.}
In contrast, the training data of the second model is formed by randomly selecting two proofs from the ten available for each theorem, irrespective of their lengths. We denote it as \textit{Model - long proof w/ t.a.e.}
For both models, we use greedy search to let them generate one tactic for each state. 
We evaluate the models on our 1000 OOD test set. 
The results are shown in the \Cref{three-model-comparison}.
A higher success rate is observed in the model trained with shorter proofs. This can be attributed to the fact that as the proof with trial-and-error information becomes longer, there is too much trial-and-error information that may detrimentally affect the model's performance, as too many failed search paths may lower the quality of the training data.

\section{Conclusion and Future Work}
In this paper, we study Automated Theorem Proving in formalized environments. We create a complete, scalable, and representative dataset of intuitionistic propositional logic theorems in Lean. We demonstrate that leveraging information from failed search states and backtracking not only teaches models how to backtrack effectively, but also helps in developing better tactics than those generated by models trained without access to backtracking insights.
We release our datasets on GitHub and Huggingface.\footnote{\propl dataset is available at \url{https://huggingface.co/datasets/KomeijiForce/PropL}. Model weights are available at \url{https://huggingface.co/KomeijiForce/llama-2-7b-propositional-logic-prover}. Generation codes for theorems and proofs are available at \url{https://github.com/ucsd-atp/PropL}. }

A natural extension of our research involves investigating whether trial-and-error information is beneficial for more general mathematical theorem-proving settings. Exploring this avenue could provide valuable insights into the effectiveness of our approach across broader mathematical domains.

\section*{Limitations}
One limitation of our study is that some proof attempts are forced to stop due to the prompt exceeding the context length of 1500 tokens. This constraint may potentially influence our results by truncating the available information during the proof search process.

Furthermore, our method was not evaluated on general mathematical theorems. This limitation arises from both the scarcity of proofs containing trial-and-error information in current math libraries and the intrinsic challenges associated with producing proofs, whether with or without backtracking, for general mathematical theorems in a formalized setting. 

Automated theorem proving with LLMs is an emerging area in machine learning. There is still a lack of baselines on LLMs to compare with our method. We establish a fundamental baseline, but we still need accumulative work to provide methods for comparison.

\section*{Ethical Consideration}

Our work learns large language models to automatically prove propositional logic theorems, which generally does not raise ethical concerns.

\section*{Acknowledgments}

This work is supported by the National Science Foundation under grants CCF-1955457 and CCF-2220892.
This work is also sponsored in part by NSF CAREER Award 2239440, NSF Proto-OKN Award 2333790, as well as generous gifts from Google, Adobe, and Teradata.

\bibliography{anthology,custom}

\newpage
\clearpage
\appendix

\section{Inference Rules for Intuitionistic Propositional Logic}
\label{sec:IRIPL}

\Cref{fig:ipl-rules} shows the inference rules for intuitionistic propositional logic. 
Rules relate sequents of the form $\Gamma \vdash A$, where $\Gamma$ is an unordered list of assumptions. A derivation is constructed from the axiom (Assumption) using the derivation rules until $\Gamma$ is empty.

For example, the ($\to$-I) rule says that from a derivation of $\Gamma \vdash B$ under the assumption of $\Gamma, A$, we can get a derivation of $\Gamma \vdash A \to B$. The ($\to$-E) rule says that from a derivation of $\Gamma \vdash A \to B$ and a derivation of $\Gamma \vdash A$ we can derive $\Gamma \vdash B$.

\begin{figure}

\begin{mathpar}
    \inferrule{ }{\Gamma, A \vdash A}\text{(Assumption)}
    
    \inferrule{ }{\Gamma\vdash T}\text{(T-I)}
    
    \inferrule{\Gamma\vdash F}{\Gamma \vdash A}\text{(F-E)}

    \inferrule{ \Gamma \vdash A \\ \Gamma \vdash B}{\Gamma \vdash A \land B}\text{($\land$-I)}

    \inferrule{ \Gamma \vdash A \land B}{\Gamma \vdash A}\text{($\land$-E1)}

    \inferrule{ \Gamma \vdash A \land B}{\Gamma \vdash B}\text{($\land$-E2)}

    \inferrule{ \Gamma \vdash A}{\Gamma \vdash A \lor B}\text{($\lor$-I1)}
    
    \inferrule{ \Gamma \vdash B}{\Gamma \vdash A \lor B}\text{($\lor$-I2)}
        
    \inferrule{\Gamma \vdash A \lor B \\ \Gamma, A \vdash C \\ \Gamma, B \vdash C}{\Gamma \vdash C}\text{($\lor$-E)}
        
    \inferrule{\Gamma, A \vdash B}{\Gamma \vdash A \to B}\text{($\to$-I)}
            
    \inferrule{\Gamma\vdash A \to B \\ \Gamma \vdash A }{\Gamma \vdash B}\text{($\to$-E)}
\end{mathpar}
    \caption{Natural Deduction Rules for Intuitionistic Propositional Logic}
    \label{fig:ipl-rules}
\end{figure}

Here are an example derivation.

$\texttt{p1} \to \texttt{p1} \lor \texttt{p2}$.

\[
\infer[\text{($\to$-I)}]{
    \vdash \texttt{p1} \to \texttt{p1} \lor \texttt{p2}
}{
    \infer[\text{($\lor$-I1)}]{
            \texttt{p1} \vdash \texttt{p1} \lor \texttt{p2}
    }{
        \infer[\text{(Assumption)}]{ 
            \texttt{p1} \vdash \texttt{p1}
        }{
        }
    }
}
\]

\section{Uniformly Distributed Data Explanation}
\label{sec:UDDE}
In this section, we discuss the uniform characteristics of our dataset, particularly emphasizing the one-to-one mapping between propositions and natural numbers. This bijection allows us to simply sample from the natural numbers to ensure the dataset exhibits uniformity.

\begin{algorithm}
\caption{Encoding}
\begin{algorithmic}[1]
\Require A tree $\mathcal{T}$ representing a proposition, with $n$ indicating the number of internal nodes and $p$ representing the number of atomic propositions
\Ensure Output a natural number

\State \Return $3^{n} (p + 2)^{n + 1}$ $\times$ \Call{ShapeNumber}{$\mathcal{T}$} $+$ \Call{AssignmentNumber}{$\mathcal{T}$}

\Function{ShapeNumber}{$\mathcal{T}$}
    \If{$\mathcal{T}$ is a single node} \Return $0$
    \EndIf
    \State $\mathcal{T}_{l}, \mathcal{T}_{r} \gets$ left and right sub-trees of $\mathcal{T}$
    \State $n_{l}, n_{r} \gets$ number of internal nodes in $\mathcal{T}_{l}, \mathcal{T}_{r}$
        \Comment{Total $n_{r} + n_{l} + 1$ internal nodes}
    \State \Return $\sum_{i = 1}^{n_{l}} C_{i - 1} C_{n_{l} + n_{r} + 1 - i}$ $+$ $C_{n_{r}}$ $\times$ \Call{ShapeNumber}{$\mathcal{T}_{l}$} $+$ \Call{ShapeNumber}{$\mathcal{T}_{r}$}
\EndFunction

\Function{AssignmentNumber}{$\mathcal{T}$}
    \State $N \gets$ \Call{NodeNumber}{root node of $\mathcal{T}$}
    \If{$\mathcal{T}$ is a single node} \Return $N$
    \EndIf
    \State $\mathcal{T}_{l}, \mathcal{T}_{r} \gets$ left and right sub-trees of $\mathcal{T}$
    \State $n_{r} \gets$ number of internal nodes in $\mathcal{T}_{r}$
    \State $A_{r} \gets 3^{n_{r}} (p + 2)^{n_{r} + 1}$
        \Comment{Compute all possible assignments in the right sub-tree}
    \State \Return $3 A_{r}$ $\times$ \Call{AssignmentNumber}{$\mathcal{T}_{l}$} $+$ $A_{r}$ $\times$ $N$ $+$ \Call{AssignmentNumber}{$\mathcal{T}_{r}$}
\EndFunction

\Function{NodeNumber}{$\mathcal{N}$}
    \State \textbf{switch} $\mathcal{N}$
    \State \quad \textbf{case} $\land$ or $T$: \textbf{return} 0
    \State \quad \textbf{case} $\lor$ or $F$: \textbf{return} 1
    \State \quad \textbf{case} $\rightarrow$: \textbf{return} 2
    \State \quad \textbf{case} $P_{i}$: \textbf{return} $i + 1$
        \Comment{$1 \leq i \leq p$}
\EndFunction
\end{algorithmic}
\label{alg:encoding}
\end{algorithm}

\subsection{Catalan Number}
The Catalan number $C_{n}$ is applicable for counting full binary trees that consist of $n + 1$ leaf nodes, corresponding to exactly $n$ internal nodes \cite{stanley2015catalan}. Additionally, it can be calculated through recursion as shown:
\begin{equation*}
    C_{n}
    = \sum_{i = 1}^{n} C_{i - 1} C_{n - i}.
\end{equation*}
The first Catalan numbers for $n = 0, 1, 2, 3, 4$ are $1, 1, 2, 5, 14$.\\
A concise interpretation of this recursion is as follows: it involves counting the number of internal nodes in the left sub-tree, amounting to $i - 1$, and then in the right sub-tree, amounting to $n - i$, for each of the $n$ scenarios in computing $C_{n}$.

\begin{algorithm}
\caption{Decoding}
\begin{algorithmic}[1]
\Require A natural number \texttt{ID}, with $n$ indicating the number of internal nodes and $p$ representing the number of atomic propositions
\Ensure Output a proposition tree
\State quotient, remainder $\gets$ \Call{DivMod}{\texttt{ID}, $3^{n} (p + 2)^{n + 1}$}
\State $S \gets$ \Call{TreeShape}{quotient, $n$}
\State $A \gets$ \Call{TreeAssignment}{$S$, remainder}
\State \Return a tree with shape $S$ and assignment $A$

\Function{TreeShape}{$N$, $n$}
    \If{$n$ is $0$} \Return a single node
    \EndIf
    \State $n_{l} \gets \max ( \{ k | \sum_{i = 1}^{k} C_{i - 1} C_{n - i} \leq N \} \cup \{ 0 \} )$
    \State $n_{r} \gets n - n_{l} - 1$
    \State remaining $\gets N - \sum_{i = 1}^{n_{l}} C_{i - 1} C_{n - i}$
    \State $N_{l}, N_{r} \gets$ \Call{DivMod}{remaining, $C_{n_{r}}$}
    \State \Return a tree with left and right sub-trees shaped by \Call{TreeShape}{$N_{l}$, $n_{l}$} and \Call{TreeShape}{$N_{r}$, $n_{r}$}, respectively
\EndFunction

\Function{TreeAssignment}{$S$, $N$}
    \State Perform an inorder traversal of the tree with shape $S$. For each leaf node, interpret it as a digit in base $p + 2$, and for each internal node, interpret it as a digit in base 3. Compute the assignment that gives $N$ by utilizing the \Call{NodeNumber}{} function introduced in \Cref{alg:encoding}
\EndFunction
\end{algorithmic}
\label{alg:decoding}
\end{algorithm}

\subsection{Bijection between Propositions and Natural Numbers}
As depicted in \Cref{fig:proposition_number_mapping}, every proposition corresponds to a unique tree representation. Consequently, it only requires the identification of a bijection between full binary trees and natural numbers.

For every full binary tree possessing $n$ internal nodes and $p$ atomic propositions, there exist
\begin{equation*}
    \begin{aligned}
        3^{\# \text{internal node}} (p + 2)^{\# \text{leaf node}}
        = 3^{n} (p + 2)^{n + 1}
    \end{aligned}
\end{equation*}
distinct cases.
Observe that the choices available for internal nodes include conjunction ($\land$), disjunction ($\lor$), and implication ($\rightarrow$); whereas for leaf nodes, the choices encompass true ($T$), false ($F$), and a set of propositions $P_{1}, \ldots, P_{p}$.

This counting facilitates an efficient ranking of all full binary trees with $n$ internal nodes.
This ranking process inherently establishes a bijection with the set of natural numbers, allowing for a clear correspondence between each full binary tree and a unique natural number.
Consequently, this sets the stage for a detailed examination of two critical processes: encoding (see \Cref{alg:encoding}), which involves mapping a proposition tree to a natural number, and decoding (see \Cref{alg:decoding}), which entails mapping a natural number back to a corresponding proposition tree.

Having established the bijection between full binary trees and natural numbers, it becomes apparent that uniformly sampling from the set of natural numbers will, in turn, result in a uniform sampling of full binary trees.

The inclusion of the parameter $n$ in \Cref{alg:encoding} is not critical for the functionality of the algorithm as $n$ can be counted from $\mathcal{T}$; however, it is retained to denote that the most basic encoding and decoding approaches are designed for a fixed number of internal nodes $n$. For example, ID 0 denotes different propositions depending on the specified value of $n$. As a result, the decoding algorithm, as outlined in \Cref{alg:decoding}, is specifically intended for uniformly generating proposition trees with a fixed $n$. To achieve a uniformly distributed proposition tree from a set of trees with various $n$, one might simply merge the rankings of fully binary trees with different $n$, a task which can be performed with relative ease. This approach of merging can be similarly applied to trees with varying $p$ as well.

Given the uncertainty surrounding the proof lengths when generating propositions, our approach involves uniform sampling for proposition selection. This selection is later refined by excluding propositions according to their proof lengths as computed by the proof generation algorithm. 


\section{Examples}
\label{sec:appendix_example}

In \Cref{fig:lean_proof}, we show an example Lean proof for a theorem in \Cref{fig:proposition_number_mapping}. Lines preceded by '--' are comments solely for explanatory purposes. 

{
\lstset{
    commentstyle=\color{gray}, 
}

\begin{figure*}[t]
\begin{lstlisting}
variable (p1 p2 p3 p4 p5 : Prop)
theorem : (((p1 ∨ p2) → False) → ((p1 → False) ∧ (p2 → False))) := by
  -- Implications on the right can always be decomposed. 
  -- Introduce an assumption h1 that says ((p1 ∨ p2) → False)
  intro h1
  -- Now we want to show ((p1 → False) ∧ (p2 → False))
  -- Conjunctions on the right can always be decomposed. 
  -- We then need to show (p1 → False) and (p2 → False) separately.
  apply And.intro
  -- We are showing (p1 → False).
  -- Implications on the right can always be decomposed.
  -- We introduce assumption h2 for p1. And we try to show False.
  intro h2
  -- We want to use the implication h1. So we show its premise.
  have h3 : (p1 ∨ p2) := by
    -- Show the left disjunct. (The right adjunct leads to an TAE)
    apply Or.inl
    -- One of the premise coincides with the conclusion.
    exact h2
  -- We have shown the premise of h1 (p1 ∨ p2), 
  -- we can now drive its conclusion (False), denoted by h4.
  let h4 := h1 h3
  -- False on the left can always be used.
  apply False.elim h4
  -- We have shown (p1 → False) and now we show (p2 → False).
  -- Implications on the right can always be decomposed.
  -- We introduce assumption h2 for p2. And we try to show False.
  intro h5
  -- We want to use the implication h1. So we show its premise.
  have h6 : (p1 ∨ p2) := by
    -- Show the right disjunct. (The left adjunct leads to an TAE)
    apply Or.inr
    -- One of the premise coincides with the conclusion.
    exact h5
  -- We have shown the premise of h1 (p1 ∨ p2),
  -- we can now drive its conclusion  (False), denoted by h7.
  let h7 := h1 h6
  -- False on the left can always be used.
  apply False.elim h7
\end{lstlisting}
\caption{An example of an intuitionistic propositional logic theorem with its proof in Lean}
\label{fig:lean_proof}
\vspace*{4in}
\end{figure*}
}

\end{document}